\documentclass[11pt,draftcls,onecolumn]{IEEEtran}

\usepackage{times}
\usepackage{epsfig}
\usepackage{graphicx}
\usepackage{amsmath}
\usepackage{amssymb}
\usepackage{algorithm}
\usepackage{algorithmic}

\begin{document}

\title{A Simple Unsupervised Color Image Segmentation Method based on MRF-MAP}

\author{Qiyang Zhao 
\thanks{Qiyang Zhao is with the School of Computer Science and Engineering, Beihang
University, Beijing, 100191, China. E-mail: zhaoqy@buaa.edu.cn.}}

\maketitle

\begin{abstract}
Color image segmentation is an important topic in the image processing field. MRF-MAP is often
adopted in the unsupervised segmentation methods, but their performance are far behind recent
interactive segmentation tools supervised by user inputs. Furthermore, the existing related
unsupervised methods also suffer from the low efficiency, and high risk of being trapped in the
local optima, because MRF-MAP is currently solved by iterative frameworks with inaccurate initial
color distribution models. To address these problems, the letter designs an efficient method to
calculate the energy functions approximately in the non-iteration style, and proposes a new binary
segmentation algorithm based on the slightly tuned Lanczos eigensolver. The experiments demonstrate
that the new algorithm achieves competitive performance compared with two state-of-art segmentation
methods.
\end{abstract}


\IEEEpeerreviewmaketitle

\section{Introduction}
\IEEEPARstart{U}{nsupervised} color image segmentation is important in various image processing and
computer vison applications, such as medical imaging \cite{Pham}, image retrieval \cite{Alajlan},
image editing \cite{Wang}, and object recognition \cite{Shotton}. Estimating the maximum a
posteriori (MAP) on the Markov random fields (MRF), is so far an fundamental tool which is widely
adopted both in the unsupervised color image segmentations and supervised ones
\cite{Kato}-\cite{Liu}. In all existing MRF-MAP-based image segmentation methods, their goals are
to find the optimal label configurations on pixels to maximize the posterior probability which is
in proportion to the product of the MRF priors and the likelyhooods terms, or equivalently, to
minimize the energy function of the smoothness terms plus data terms.

The likelyhood/data terms consist of the parameters of some color distribution models, here these
models specify the probabilities of any color occurring in each segmentation. These parameters
usually derive from user interactions or random sampling, hereby there is always non-negligible
inaccuracy in the likelyhood terms. In order to address this, the Expectation-Maximization (EM),
simulated annealing and other iterative methods are usually adopted to progressively approach the
appropriate parameters, especially in the case of unsupervised segmentations up to today
\cite{Kato}-\cite{Celeux}. There are many choices of optimization algorithms to be adopted in the
M-step \cite{Szeliski}.

There are three major disadvantages in the current unsupervised segmentation methods. The first is
the low efficiency of the iterative frameworks, particularly when faced with large size images. The
second is always on the high risk to be trapped to the local optima of the energy function.
Although stepping out and restarting the iteration is a reasonable improvement, there would
inevitably be additional computational load and it is possible to be trapped again. The third is
the coarseness in the segmentation results, and it is partially caused by the roughness of the
likelyhood parameters. To address these issues, the letter proposes a new unsupervised binary
segmentation method based on the approximation of the likelyhood terms, where the iterative
computations are replaced by only one single step of solving the eigenvector of the largest
eigenvalue, therefore the computational efficiency is remarkably improved. This new approach
increases the chance to high quality segmentation results by obtaining the nearly optimal solutions
to maximize the posterior probability as possible. It also provides us an effective way to test and
verify the MRF prior parameters or their involved generating schemes, which are also critical to
the segmentation tasks.

\section{Color Image Segmentation based on MRF-MAP}

In the following sections, we focus on the binary segmentation with the label set \{\textit{fore,
back}\}. The computational goal of these methods is to maximize the probability $P(L|I)$ of the
segmentation label configuration $L$ given an image $I$. According to the Bayesian rule, it is
equivalent to maximize the joint probability $P(I, L)=P(I|L) \cdot P(L)$, where the prior $P(L)$ is
established on the Markov random field of $I$, the conditional probability $P(I|L)$ is the
likelyhood that the pixel colors occur in their corresponding segments marked by different labels.
In a more prevailing view, what we need is to minimize a energy function $E $ which is the negative
log-hood of $P(I, L)$. Here $E$ is usually written in the form of a data term $E_{D}$ plus a
smoothness term $ E_{S}$ multiplied by a factor $\lambda$:
\begin{equation}
E=E_{D}+\lambda E_{S}
\end{equation}
where $E_{D}$ reflects the likelyhood of the color occurrences in the image segments, $E_{S}$ is
the sum of all adjacency interaction potentials of each two neighboring pixels of different labels:
\begin{equation}
E_{D}=\sum\limits_p{-\ln
P_{L(p)}(p)},E_{S}=\!\!\!\!\!\!\!\!\sum_{\mbox{\scriptsize$\begin{array}{c}
{(p,q)\in N}\\
{L(p)\ne L(q)}\end{array}$}}\!\!\!\!\!\!\!\!\!\!{S(c(p),c(q))}
\end{equation}
where $L(\cdot)$ is the pixel label, $c(\cdot)$ is the pixel color, and $S(\cdot,\cdot)$ is the
perceptually similarity weight of two colors. It is meant two pixels $p$ and $q$ are adjacent to
each other by noting $(p,q)\in N$. Thereafter, the segmentation task is to pursue an appropriate a
label configuration to reach the lowest energy. Although as indispensable as the data terms when
computing the energy functions, the smoothness terms are not to be addressed in the letter.

There are two steps when determining the coefficients in the data term $E_{D}$. First, a suitable
color statistical model should be chosen. Histograms are usually adopted for images of small color
spaces, such as gray scale or 256 colors, but it is not suitable for large color spaces as the
samples were statistically too few when facing so many histogram bins. Some other  models  fit
large color spaces well, and make an appropriate comprise between the efficiency and accuracy, such
as the Gaussian mixture model (GMM). Second, the model parameters should be determined. However
here arises a \textit{chicken or the egg} dilemma unavoidably: we have to know the parameters first
to minimize the energy function to obtain the optimal segmentation, but the optimal segmentation is
just the key to produce the accurate parameters mentioned above. The usually adopted solutions to
this are the iterated procedures, such as EM, in which the estimation and optimization are
performed sequently but isolatedly in each single loop. Here the initial parameters are determined
from sample pixels chosen by user interactions or random samplings. There are many choices to
perform the optimization \cite{Szeliski}: graph cut, Loopy Belief Propagation (LBP) and Iterated
Conditional Model (ICM).

The low computational efficiency is an adherent shortcoming of the iterated solutions, as it is
extremely hard to predict when and where the iterations would halt. Furthermore, although the
minimum of the energy function is an unambiguous target itself, the actual aim of the iterated
solutions is not mathematically explicit for us to approach. As a result, the iterations are likely
finished at the local minimal in most cases. To address these issues, the letter proposes an
approximating expression which is rather close to $E_{D}$ in (2), and associate the approximated
target energy function with the cut on a complete graph $G$ which has both positive and negative
edge weights. In the following manipulations, it is rather straightforward to solve an eigen-system
to pursue the minimum cut $C$ on $G$, so to minimize the energy. There the expected segmentations
are worked out directly without considering the troublesome parameters of the data terms at all.

\section{Segmentation based on Approximate MRF-MAP}

Consider there are $n$ pixels of $m(m\!\!<\!<\!\!n)$ colors in the image $I$, and any two colors
are perceptually distinguishable from each other. It is almost always achievable with the help of
the existing color clustering algorithms, even facing much splendid images. Therefore we choose the
histograms as our color distribution models in (2). Let $c(p)$ be the color of the pixel $p$. For
all pixels having color $i$, let $n_{i}$ be the total amount, $n_{f,i}$ be the amount of those
having label \textit{fore}, so be $n_{b,i}$ and label \textit{back}. Clearly
$n_{i}=n_{f,i}+n_{b,i}$. Let $F$ and $B$ be the two sets of \textit{fore} and \textit{back} pixels
respectively, and the corresponding pixel amounts are $n_{f}=n_{f,1}+n_{f,2}+\ldots+n_{f,m}$ and
$n_{b}=n_{b,1}+n_{b,2}+\ldots+n_{b,m}$. Leaving the smoothness term $E_{S}$ unchanged, we have the
the data term $E_{D}$ as
\begin{equation}
\begin{split}
 E_{D}&=\sum\limits_{p\in F} {-\ln \frac{n_{f,c(p)} }{n_{f}}} +\sum\limits_{p\in
B} {-\ln \frac{n_{b,c(p)} }{n_{b} }} \\
 &=(n_f \!\ln\! n_f + n_b \!\ln\! n_b)\!-\!\sum\limits_{i=1}^m ({n_{f,i} \!\ln\!
 n_{f,i} + n_{b,i} \!\ln\! n_{b,i}}) \\
 &=[n \cdot (\frac{n_f}{n} \!\ln\! \frac{n_f}{n} + \frac{n_b}{n} \!\ln\! \frac{n_b}{n})+n\ln n] \\
 &\ \ \ -\!\sum\limits_{i=1}^m [n_i \cdot ({\frac{n_{f,i}}{n_i} \!\ln\!
 \frac{n_{f,i}}{n_i} + \frac{n_{b,i}}{n_i} \!\ln\! \frac{n_{b,i}}{n_i}})+n_i \ln n_i]
\end{split}
\end{equation}
Now consider a function $g(x)$ defined on the interval $[0, 1]$:
\begin{equation}
g(x)=\begin{cases}x\ln x+(1-x)\ln(1-x), \mbox{if}\ 0 < x < 1. \\
0, \mbox{otherwise.} \\
\end{cases}
\end{equation}
Clearly $g$ is continuous on the whole interval $[0, 1]$. Expanding the log terms into the Taylor
series and simplifying the expression, we have $g(x)=-\frac{5}{2}xy+\Delta (x)$, where $y=(1-x)$
and
\begin{equation}
\Delta (x)=-xy[\frac{1}{3}(x^2+y^2)+\frac{1}{4}(x^3+y^3)+\cdots],
\end{equation}
which has the mean value $\int_0^1 {\Delta (x)dx=-\frac{1}{12}}$ on [0, 1]. If we replace
\textit{$\Delta $}($x)$ with the constant $-\frac{1}{12}$ on the entire interval, the mean squared
error would be $\int_0^1 {(\Delta (x)-\frac{1}{12})^2dx< }\ 3\times 10^{-4}$. Since the MSE is
considerable small, it is totally acceptable to approximate $g$ with $g^\ast=
{-\frac{5}{2}xy-\frac{1}{12}}$, as shown in Fig. 1.
\begin{figure}[htbp]
\centerline{\includegraphics[width=3.2in,height=1.8in]{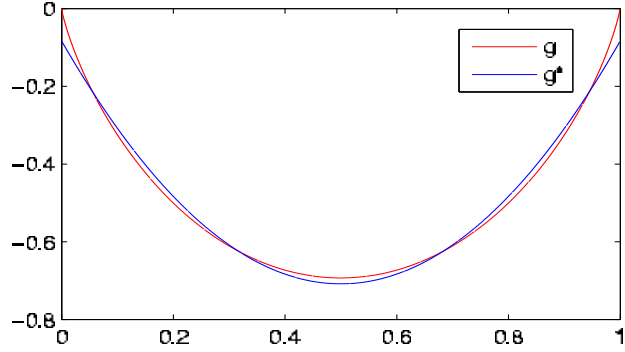}} \caption{Comparison of function $g$
and its approximation $g^\ast$ on [0, 1].}
\end{figure}

With the help of $g^\ast$, now we can approximate the energy function $E$ in (3) with
\begin{equation}
\begin{split}
&({-\frac{5}{2n}n_f\!\cdot\!n_b\!-\!\frac{1}{12}}\!+\!n\!\ln\!n)\!-\!\sum\limits_{i=1}^m {(
{-\frac{5}{2n_i
}n_{f,i}\!\cdot\!n_{b,i}\!-\!\frac{1}{12}\!+\!n_i\!\ln\!n_i})} \\
&+\lambda\cdot\!\!\!\!\!\!\!\!\sum_{\mbox{\scriptsize$\begin{array}{c}
{(p,q)\in N}\\
{L(p)\ne L(q)}\end{array}$}}\!\!\!\!\!\!\!\!\!\!{S(c(p),c(q))}
\end{split}
\end{equation}
Since $m$, $n$ and all $n_{i}$'s are constant in the input image, it is actually to minimize the
following function $E^\ast$ when minimizing the above expression as the approximation of $E$:
\begin{equation}
\begin{split}
E^\ast\!\!=\!\!-\frac{5}{2n}n_f\!\cdot\!n_b\!\!+\!\!\sum\limits_{i=1}^m
{\!({\frac{5}{2n_i}n_{f,i}\!\cdot\!n_{b,i}}
)}\!+\!\lambda\!\cdot\!\!\!\!\!\!\!\!\!\!\!\!\sum_{\mbox{\scriptsize$\begin{array}{c}
{(p,q)\in N}\\
{L(p)\ne L(q)}\end{array}$}}\!\!\!\!\!\!\!\!\!\!\!\!{S(c(p),c(q))}.
\end{split}
\end{equation}

Now construct an undirected complete graph $G$ of $n$ nodes each of which corresponds to one pixel,
and set the edge weight $w(p, q)$ to be the sum of the following three terms
\begin{equation}
\begin{split}
&w_1 (p,q)=-\frac{5}{2n} \\
&w_2 (p,q)=\left\{ {\begin{array}{l}
 \frac{5}{2n_i },\mbox{ if }c(p)=c(q)=i, \\
 0,\mbox{ otherwise.} \\
 \end{array}} \right.\\
&w_3 (p,q)=\left\{ {\begin{array}{l}
 \lambda \cdot S(c(p),c(q)),\mbox{ if }{(p,q)\in N}, \\
 0,\mbox{ otherwise.} \\
 \end{array}} \right.
\end{split}
\end{equation}
It is easy to prove that, for any label configuration $L$ of the binary segmentations, $E^\ast$ is
equal to the capacity of the  cut $C= \{F, B\}$ arising from $L$ on $G$. Therefore it is equivalent
to find the minimum cut on $G$ when minimizing $E^\ast$. However the existing minimum cut
algorithms are not suitable here because of the existence of negative edge weights. In fact, the
problem here is computationally equivalent to a well-known NP-complete problem, Max-Cut, on the
graphs with non-negative weights. Therefore it is hard to obtain the exact minimum of $E^\ast$ in
polynomial time.

Our solution to this is to generalize it into the continuous real space $R^n$. First we put the
label configuration into an indicator vector $D=\left[ {d_1 ,d_2 ,\cdots ,d_n } \right]^T$: for
$i=1,2,\cdots\,n$, let $d_i=+1$ if the $i$th pixel label is \textit{fore}, and $-1$ for
\textit{back}. Then establish a matrix $W=[w(p,q)]$ and let $S_{W}$ denote the sum of all its
entries. It is easy to prove that the cut value is equal to $\frac{1}{2}({S_W -D^TWD})$. After
generalizing the $d_i$'s to be in the continuous interval $[-1, 1]$ instead of $\{+1,-1\}$, our
task becomes into
\begin{equation}
\max D^TWD,\mbox{s.t.}\left\| D \right\|_2 =n
\end{equation}
because $S_W$ is a constant here. According to the \textit{Lagrange multiplier} method
\cite{Golub}, the solution to (9) is the eigenvector $D^\ast$ corresponding to the largest
eigenvalue of $W$. The \textit{Lanczos} algorithm, well known as the fastest method solving
extremal eigenvectors for large sparse matrices, is adopted here to calculate $D^\ast=[{d^\ast_1
,d^\ast_2,\cdots,d^\ast_n}]^T$. Since $W$ is full, the embedded matrix-vector multiplication must
be improved using the special structure of $W$. At last, we get the required binary labels
straightforwardly by setting the $i$th label to be $fore$ if $d^\ast_i \ge 0$, or $back$ if
$d^\ast_i < 0$. Here is the outline of our new segmentation algorithm:

\begin{algorithm}
\caption{Color image segmentation based on MRF-MAP}
\begin{algorithmic}[1]
\STATE Clustering all colors into $m$ classes; compute $S(c(p),c(q))$ for each pair of adjacent
pixels $p$ and $q$;

\STATE Calculate the largest eigenvector $D^\ast=\left[ {d^\ast_1 ,d^\ast_2 ,\cdots ,d^\ast_n }
\right]^T$ of the matrix $W$ with \textit{Lanczos} eigensolver, where we obtain the product
$R=[r_1,r_2,\cdots,r_n]^T$ of $W$ and any vector $V=[{v_1,v_2,\cdots,v_n }]^T$ as

\begin{itemize}
\item let \textit{$\varphi $} = 0, and \textit{$\theta $}$_{i}$ = 0 for all $i=1,2,\cdots,m$;

\item for $k=1$ to $n$, let $\varphi\leftarrow\varphi-\frac{5}{2n}\cdot v_k$,
$\theta_{c(k)}\leftarrow\theta_{c(k)}+\frac{5}{2n_{c(k)} }\cdot v_k $, and
$\mu_k=\sum\nolimits_{j:(j,k)\in N}{S(j,k)\cdot v_j}$;

\item for $k=1$ to $n$, $r_k \leftarrow \varphi + \theta_{c(k)} + (\frac{5}{2n}-\frac{5}{2n_{c(k)}})\cdot v_k$.

\end{itemize}

\STATE Output the label of the $k$th pixel as \textit{back} if $d^\ast_k < 0$, or \textit{fore} if
$d^\ast_k \ge 0$.

\end{algorithmic}
\end{algorithm}

The analysis on the computational complexity is rather straightforward: the step 1, 3 can be
finished in $O(n)$ time; the matrix-vector multiplication can also be finished in $O(n)$ time, so
the \textit{Lanczos} algorithm revoked in step 2 can be finished in $O(n\cdot d)$ time, where $d$
is the amount of performed iterations solving the eigenvector. Therefore the total time complexity
of is $O(n\cdot d)$. Since $d$ is irrelevant with $n$ and  empirically always less than a certain
constant, the new segmentation algorithm is practically a nearly linear one.

\section{Experiments}
The experiments are finished on color images chosen from two segmentation datasets from Berkeley
and MS research at Cambridge, together with the source codes developed in Matlab. The color
clustering method adopted in step 1 is from \cite{Orchard}, and all of the amounts of color classes
are set to be 16 for these images. $\lambda$ varies from 1 to 10, and all smoothness terms are
simply set to be 1 over all adjacent pixels despite their colors, to say, $S(\cdot,\cdot)\equiv 1$.

\begin{figure*}
\centerline{\includegraphics[width=6.5in,height=5.2in]{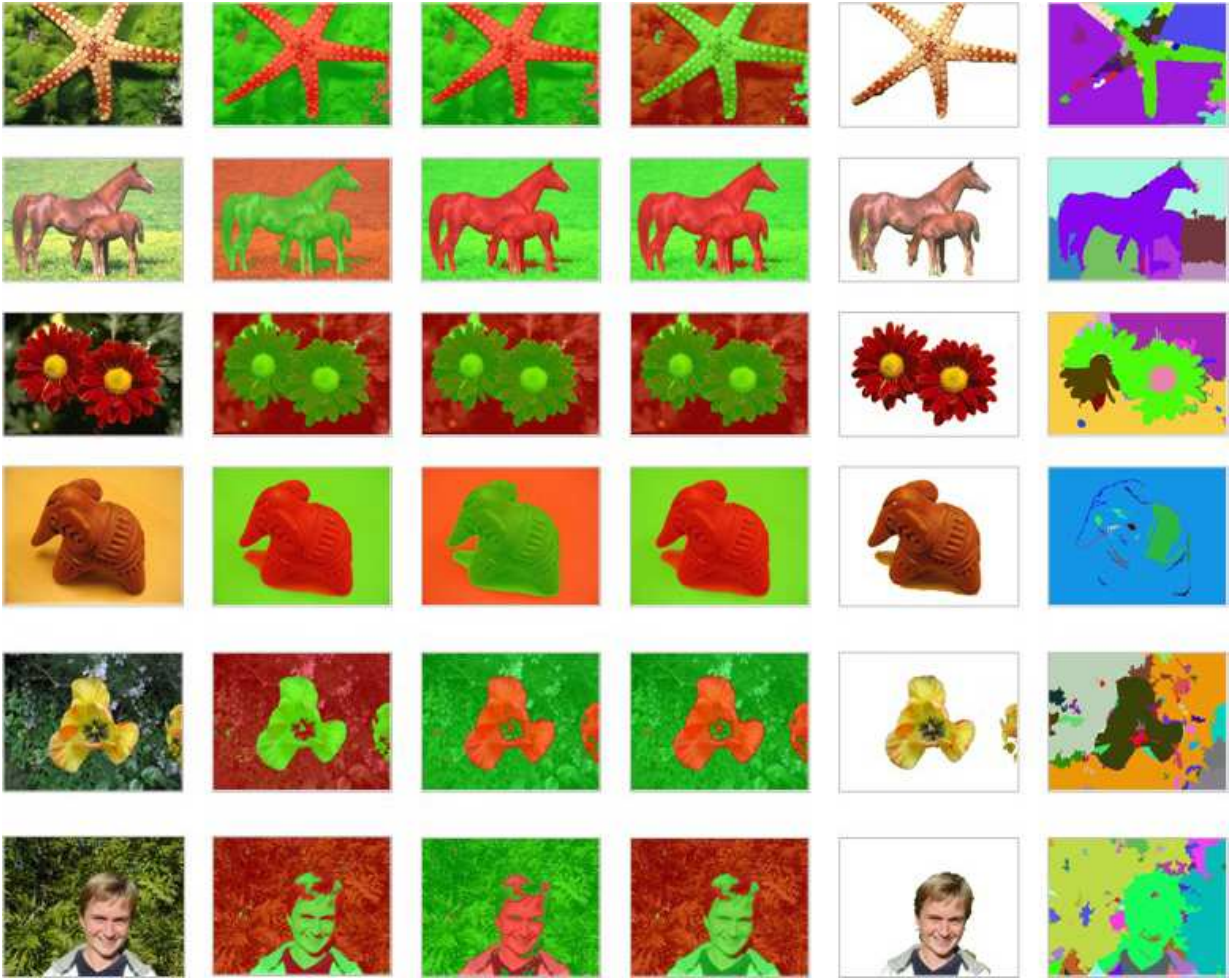}} \caption{Comparison of the
segmentation results by our method and other two methods. (Leftmost: original images. Other columns
from left to right: segmentation results by our method with $\lambda=1,5,10$, by the method in
\cite{Rother} and \cite{Felzenszwalb}.)}
\end{figure*}

These images are intentionally chosen to be of splendid colors and delicate local details, so to
verify the performance of our new method facing different challenges. Two state-of-art segmentation
algorithms, one supervised \cite{Rother} and another unsupervised \cite{Felzenszwalb}, are chosen
in the control experiments to examine the segmentation quality of our new method. In general, all
segmentation results of our new method are basically acceptable in the experiments. Especially, the
perceptually outstanding objects, if any in the test images, are usually figured out of the
underlaying scenes accurately. There are six groups of typical experiment results in Fig. 2: each
group includes five different segmentation results of an identical color image, three of them are
all from our method but with different $\lambda$'s, whereas the rest two are from the other two
methods as the comparison. The embedded objects, striking either for fresh colors or large
continuous shapes, are precisely outlined by our new method in Fig. 2. Its segmentation quality, is
considerably close to the user-interactive-styled method in \cite{Rother}, but much better than the
unsupervised one in \cite{Felzenszwalb}.

Consequentially, there are some noticeable slight differences between the segmentation results on
different $\lambda$'s of our new method. There are more isolated, but vivid pieces when
$\lambda=1$, and simultaneously the segmentation boundaries are more likely located on the desired
edges of the objects. However, it also brings too much emphasis on these discontinuous line
segments, and results in much more isolated pieces in the segmentations. When $\lambda$ varies from
1 to 5, then to 10, it is shown the segmentation boundaries become smoother and smoother on the
cost of losing the elaborate details, and the two segmentation zones are more close to each other
in sizes when $\lambda=10$. The reason for that is, the smoothness terms become larger and larger
quantitatively so that the continuity of the segmentations is emphasized much more, therefore it is
more inclined to cut two pieces of the the same sizes and flat boundaries to reach the minimum
energies.

\section{Conclusions}
In this letter, a new unsupervised MRF-MAP-based segmentation algorithm is introduced. By
introducing an reasonable approximation to the data terms, the energy functions could be minimized
remarkably, however, without any supervision. The new method is able to obtain the high-quality
segmentation results, as well as the high computational efficiency. The future work includes the
investigating the computational hardness of MRF-MAP, and extending the new method to video
segmentations and multiple-labeled segmentations.

\end{document}